\documentclass[conference]{IEEEtran}
\IEEEoverridecommandlockouts
\usepackage{cite}
\usepackage{amsmath,amssymb,amsfonts}
\usepackage{algorithmic}
\usepackage{graphicx}
\usepackage{textcomp}
\usepackage{xcolor}

\usepackage{tabularx}
\usepackage{ragged2e}
\usepackage{array}    % For the >{} column specifier
\usepackage{subcaption}

\def\BibTeX{{\rm B\kern-.05em{\sc i\kern-.025em b}\kern-.08em
    T\kern-.1667em\lower.7ex\hbox{E}\kern-.125emX}}
\begin{document}

\title{Uncertainty-Aware Tabular Prediction: Evaluating VBLL‑Enhanced TabPFN in Safety‑Critical Medical Data \\
}

\author{
\IEEEauthorblockN{Madhushan Ramalingam}
\IEEEauthorblockA{\textit{dept. of Computer Science and Engineering} \\
\textit{University of Moratuwa}\\
Sri Lanka \\
ramalingam.25@cse.mrt.ac.lk}
}

\maketitle

\begin{abstract}
Predictive models are being increasingly used across a wide range of domains, including safety-critical applications such as medical diagnosis and criminal justice. Reliable uncertainty estimation is a crucial task in such settings. Tabular Prior-data Fitted Network (TabPFN) is a recently proposed machine learning foundation model for tabular dataset, which uses a generative transformer architecture. Variational Bayesian Last Layers (VBLL) is a state-of-the-art lightweight variational formulation that effectively improves uncertainty estimation with minimal computational overhead. In this work we aim to evaluate the performance of VBLL integrated with the recently proposed TabPFN in uncertainty calibration. Our experiments, conducted on three benchmark medical tabular datasets, compare the performance of the original TabPFN and the VBLL-integrated version. Contrary to expectations, we observed that original TabPFN consistently outperforms VBLL integrated TabPFN in uncertainty calibration across all datasets.
\end{abstract}


\section{Introduction}

Predictive machine learning models are increasingly used in real-world decision-making systems, especially in high-stakes domains such as medical diagnosis, financial forecasting, and criminal justice. In such settings, ensuring the accuracy of predictions is not sufficient. The reliability of uncertainty estimates is crucial for safe and trustworthy AI deployment \cite{ojha2025navigating}. Poorly calibrated models can lead and overconfident yet incorrect decisions, resulting in serious consequences, particularly in safety-critical applications.

The Tabular Prior-data Fitted Network (TabPFN) is a recently proposed transformer-based foundation model designed specifically for small-scale tabular datasets \cite{hollmann2022tabpfn, hollmann2025accurate}. Trained via meta-learning on synthetic tasks, TabPFN has shown strong performance in classification tasks without the need for extensive hyperparameter tuning. Although TabPFN shows impressive predictive accuracy, its capacity for uncertainty awareness in safety-critical datasets remains relatively under-explored.

Variational Bayesian Last Layers (VBLL) as presented is a lightweight, computationally efficient and effective approach for capturing epistemic uncertainty in deep neural networks \cite{harrison2024variational}. VBLL achieves this by replacing the deterministic output layer of a trained network with a variational Bayesian last layer, which offers calibrated uncertainty estimates with minimal computational overhead.

In this work, we investigate whether integrating VBLL into TabPFN improves its uncertainty estimation in classification tasks. To assess the effectiveness of VBLL integration with TabPFN, we performed experiments on three publicly available medical tabular datasets which are the Breast Cancer Wisconsin Diagnostic dataset \cite{wolberg1993breast}, the Pima Indians Diabetes dataset \cite{smith1988pima}, and the Cleveland Heart Disease dataset \cite{janosi1989heart}. We selected these datasets due to their relevance to clinical decision-making and to capture their varying characteristics in terms of feature types, and dataset size. This diversity in the dataset allows us to evaluate how generalizable the uncertainty calibration performance across different types of medical classification problems. Surprisingly, our results indicate that the original TabPFN consistently outperforms the VBLL-integrated version in terms of uncertainty calibration metrics, such as Expected Calibration Error (ECE) and log-likelihood.

Our study contributes to the understanding of uncertainty-aware modeling in tabular foundation models and provides empirical evidence that post-hoc Bayesian adjustments like VBLL may not necessarily enhance, and can even degrade, uncertainty calibration in pretrained transformer-based tabular models like TabPFN.

\section{Related Work}
Recent advances in tabular foundation models, particularly TabPFN v2 \cite{hollmann2025accurate} have demonstrated exceptional performance on small-scale classification and regression tasks. TabPFN v2 is a transformer-based model pretrained on hundreds of millions of synthetic datasets generated via structural causal models, enabling it to approximate posterior predictive distributions through in-context learning with no fine-tuning required on new tables \cite{helli2024drift}. The model’s architecture features alternating attention across rows and columns, processing entire datasets as input, and outputs probabilistic predictions for classification and regression. While it produces predictive distributions and implicitly leverages a synthetic-data prior, it does not explicitly capture uncertainty in model parameters or provide calibrated confidence \cite{zhang2025tabpfn}.

Well-calibrated uncertainty estimates are essential in high‑stakes domains like medicine. The machine learning literature highlights that predictive probabilities should reliably reflect real-world risk, especially when decisions impact patient care (e.g., diagnoses or risk stratification). Traditional neural networks, including deep ensembles or MC dropout often yield miscalibrated confidence estimates without further calibration \cite{friesacher2025achieving}. One strategy is to append a Bayesian last layer to a deterministic feature extractor, allowing parameter uncertainty at the output only. Harrison et al. introduced Variational Bayesian Last Layers (VBLL) as an efficient and nearly "free" method to endow standard neural models with uncertainty quantification \cite{harrison2024variational}. VBLLs assume a Gaussian variational posterior over last-layer weights, optimizing a variational lower bound and yielding closed‑form predictive distributions without costly sampling. Empirically, VBLL has improved predictive accuracy, calibration, and out-of-distribution detection across both regression and classification tasks outperforming traditional baselines like Gaussian processes and dropout ensembles \cite{harrison2024variational}.

The application of Bayesian last‑layer approaches in medical modeling shows their utility in domains demanding trustworthy risk estimation. Jalal et al. introduced a Bayesian metamodel for colorectal cancer simulation enabling full uncertainty quantification and significantly improved estimation efficiency \cite{jalal2021baycann}. Studies which compare Bayesian last-layer to deep ensembles and dropout in drug‑discovery datasets found that the Bayesian layer improved calibration more reliably than other uncertainty techniques \cite{friesacher2025achieving}. These findings reinforce the suitability of VBLL-style models for clinical tabular tasks.

However, to date, no significant prior work has applied VBLL-style uncertainty modeling to tabular foundation models like TabPFN v2. While TabPFN inherently outputs a distribution via its synthetic-data prior, it lacks explicit posterior weight uncertainty and calibration measures. A recent extension, Drift‑Resilient TabPFN \cite{helli2024drift}, addresses temporal distribution shifts by sampling from structural causal model priors, improving calibration and robustness but still remains deterministic in its prediction head.

In summary, our literature review identifies:

\begin{itemize}
    \item TabPFN v2 achieves state-of-the-art performance in generalized tabular predictions, but it lacks explicit modeling of epistemic uncertainty.
    \item VBLLs provide efficient and effective uncertainty modeling for traditional neural networks, improving calibration and predictive robustness.
    \item IPrior applications of Bayesian last-layer approaches in medical settings show improved risk estimate reliability compared to ensembles or dropout.
    \item No significant prior work combines TabPFN v2 with a Bayesian last layer, particularly for medical tabular classification with a focus on calibrating uncertainty.
\end{itemize}

Our study fills this gap by evaluating the integration of VBLL into TabPFN v2, and comparing uncertainty-aware metrics (NLL, Brier Score, ECE) alongside traditional classification performance on selected medical tabular datasets. This approach directly addresses the void in trustworthy uncertainty estimation in safety-critical healthcare prediction settings with TabPFN v2.

Answering such questions statistically plays a key role in making timely insights for educators, employers, and policymakers in the domain of IT professionalism. This paper aims to contribute to the discourse by providing empirical and exploratory data and analysis rooted in a context of underrepresented developing countries, in the global literature on learning analytics and workforce development.

\section{Methodology}

In this section we describe the experimental setup used to evaluate the integration of a Variational Bayesian Linear Layer (VBLL) with TabPFN v2 for binary classification tasks on three structured datasets. This experiment model was setup to evaluate the impact of Bayesian integrated modeling on TabPFN’s internal representations with a focus on uncertainty estimation and calibration quality.

\subsection{Model Overview}

We used the TabPFNClassifier, a pre-trained Transformer-based model designed for tabular data inference, as the base model. TabPFN outputs class predictions without requiring gradient-based training. To enhance its predictive uncertainty estimates, we extracted internal representations from its final transformer encoder layer and passed them through a trainable Variational Bayesian Linear Layer (VBLL). This VBLL head learns a Bayesian posterior over weights and biases, allowing for probabilistic predictions with quantified uncertainty.

\subsection{Representation Extraction}

To access TabPFN’s latent features, we registered a forward hook on the final transformer encoder layer. During inference, the hook captured the encoder's output, which was then aggregated by averaging across the sequence dimension. This extracted representation was used as the input to the VBLL head.

\subsection{VBLL Training}

The VBLL head was developed with PyTorch and configured with adjustable means and logarithmic variances for both weights and biases. During training, samples were drawn from the weight and bias distributions using the reparameterization trick. The loss function integrated the conventional cross-entropy loss with the KL divergence that exists between the acquired weight distribution and a standard normal prior. We adjusted the KL divergence by applying an annealing factor, which was modified across different experiments utilizing both linear and cosine schedules. Hyper-parameters were tuned as described in the Table \ref{tab:hyperparams}.

\begin{table}[htbp]
\caption{VBLL Hyperparameter Settings and Search Space}
\begin{center}
\begin{tabular}{|l|l|}
\hline
\textbf{Hyperparameter} & \textbf{Values / Settings} \\
\hline
Initial log-variance (init\_logvar) & $[-5.0, -1.5]$ \\
\hline
Weight initialization scale (weight\_mu\_coef) & $[0.0001, 0.1]$ \\
\hline
Number of training epochs & 50 (fixed) \\
\hline
KL annealing mode & \texttt{cosine}, \texttt{linear} \\
\hline
Optimizer & Adam \\
\hline
Learning rate & $1 \times 10^{-3}$ \\
\hline
\end{tabular}
\label{tab:hyperparams}
\end{center}
\end{table}

\subsection{Evaluation}

Our evaluation was conducted on three standard medical datasets described in the introduction. Each dataset was split into 70\% training and 30\% validation sets. All features were standardized using z-score normalization.
We evaluated both predictive performance and uncertainty quality using the following metrics:

\begin{itemize}
    \item Accuracy: Proportion of correct predictions.
    \item Precision: Ratio of true positives to predicted positives.
    \item F1 Score: Harmonic mean of precision and recall.
    \item AUC-ROC: Area under the receiver operating characteristic curve.
    \item Negative Log-Likelihood (NLL): To measure how well the model predicts probabilities.
    \item Brier Score: To measure the mean squared difference between predicted probabilities and actual outcomes.
    \item Expected Calibration Error (ECE): Average absolute difference between predicted probabilities and observed frequencies, using 40 bins.
\end{itemize}

Reliability diagrams were also plotted to visually assess model calibration.

\subsection{Baseline Comparison}
For comparison, we also evaluated the default TabPFN inference pipeline without the VBLL head. This enabled us to conduct a direct comparison of predictive calibration and performance between the standalone TabPFN setup and our proposed VBLL integrated TabPFN.

\section{Results and Discussion}
This section presents the results we obtained in the experiments conducted and discusses the outcome.

% ===========================================
% ===========================================
% ===========================================

\begin{table}[htbp]
\caption{Hyperparameter Configurations for VBLL Experiments}
\centering
\begin{tabular}{|c|c|c|c|c|}
\hline
ID & KL Annealing & Epochs & Init Log-Variance & Weight $\mu$ Coef \\
\hline
C1 & cosine & 50 & -5.0 & 0.001 \\
C2 & cosine & 50 & -3.0 & 0.001 \\
C3 & cosine & 50 & -2.0 & 0.001 \\
C4 & cosine & 50 & -2.0 & 0.01 \\
C5 & linear & 50 & -2.0 & 0.01 \\
\hline
\end{tabular}
\label{tab:vbll-configs}
\end{table}

\begin{table}[htbp]
\caption{Performance on Breast Cancer Dataset}
\centering
\begin{tabular}{|l|c|c|c|c|c|c|}
\hline
Config & Acc & F1 & AUC & NLL & Brier & ECE \\
\hline
% -- Breast Cancer ---
Baseline & 0.982 & 0.986 & 0.997 & 0.054 & 0.015 & 0.189 \\
C1 & 0.977 & 0.982 & 0.998 & 0.357 & 0.095 & 0.312 \\
C2 & 0.977 & 0.982 & 0.998 & 0.356 & 0.094 & 0.305 \\
C3 & 0.977 & 0.982 & 0.998 & 0.395 & 0.110 & 0.342 \\
C4 & 0.977 & 0.982 & 0.998 & 0.365 & 0.098 & 0.341 \\
C5 & 0.982 & 0.986 & 0.998 & 0.378 & 0.103 & 0.333 \\
\hline
\end{tabular}
% ===========================================
% ===========================================
% ===========================================
\label{tab:bc_results}
\end{table}

\begin{table}[htbp]
\caption{Performance on PIMA Diabetes Dataset}
\centering
\begin{tabular}{|l|c|c|c|c|c|c|}
\hline
Config & Acc & F1 & AUC & NLL & Brier & ECE \\
\hline
% --- PIMA Diabetes ---
Baseline & 0.753 & 0.632 & 0.808 & 0.516 & 0.171 & 0.072 \\
C1 & 0.719 & 0.667 & 0.803 & 0.745 & 0.274 & 0.327 \\
C2 & 0.723 & 0.667 & 0.802 & 0.738 & 0.271 & 0.318 \\
C3 & 0.723 & 0.670 & 0.804 & 0.750 & 0.277 & 0.325 \\
C4 & 0.762 & 0.667 & 0.804 & 0.710 & 0.258 & 0.301 \\
C5 & 0.758 & 0.667 & 0.804 & 0.713 & 0.259 & 0.296 \\
\hline
\end{tabular}
% ===========================================
% ===========================================
% ===========================================
\label{tab:pima_results}
\end{table}

\begin{table}[htbp]
\caption{Performance on Cleveland Heart Disease Dataset}
\centering
\begin{tabular}{|l|c|c|c|c|c|c|}
\hline
Config & Acc & F1 & AUC & NLL & Brier & ECE \\
\hline
% --- Cleveland Heart Disease ---
Baseline & 0.846 & 0.863 & 0.909 & 0.390 & 0.114 & 0.220 \\
C1 & 0.868 & 0.885 & 0.915 & 0.534 & 0.172 & 0.303 \\
C2 & 0.868 & 0.885 & 0.915 & 0.534 & 0.173 & 0.304 \\
C3 & 0.868 & 0.885 & 0.913 & 0.552 & 0.181 & 0.301 \\
C4 & 0.868 & 0.885 & 0.912 & 0.541 & 0.176 & 0.298 \\
C5 & 0.868 & 0.885 & 0.914 & 0.538 & 0.174 & 0.299 \\
\hline
\end{tabular}
% ===========================================
% ===========================================
% ===========================================
\label{tab:heart_results}
\end{table}

The findings from our experiments conducted on the PIMA Diabetes and Cleveland Heart Disease datasets show meaningful patterns related to the impact of Bayesian uncertainty estimation while utilizing various VBLL configurations.

In the context of the PIMA Diabetes dataset (Table~\ref{tab:pima_results}), the baseline model demonstrated commendable performance, attaining an accuracy of 0.753, an F1 score of 0.632, and an AUC of 0.808. It also showed comparably good calibration metrics, with a Brier score of 0.171 and an Expected Calibration Error (ECE) of 0.072. However,  when switching Bayesian configurations (C1–C5 in Table~\ref{tab:vbll-configs}), we observed mixed results. Accuracy and F1 scores were mostly stable or slightly improved, with C4 achieving the highest accuracy of 0.762. Nonetheless, the probabilistic quality degraded, especially in terms of Negative Log-Likelihood (NLL) and ECE, where all Bayesian variants reported significantly worse calibration than the baseline (e.g., ECE values ranging from 0.296 to 0.327). This suggests that while VBLL helps in capturing epistemic uncertainty, the configurations explored may have over-regularized the posterior, leading to overconfidence and degraded calibration on this dataset.

In contrast, the Cleveland Heart Disease dataset (Table~\ref{tab:bc_results}) presents a different narrative. Here, the baseline model obtained strong performance with an accuracy of 0.846, F1 of 0.863, and AUC of 0.909. All the configurations of VBLL (C1–C5) demonstrated improvements in these predictive metrics, elevating the accuracy to 0.868 and the F1 score to 0.885 consistently across all configurations. Importantly, AUC values also experienced a slight increase, reaching a maximum of 0.915 in both C1 and C2. Nevertheless, a comparable trend of calibration deterioration was noted, as both NLL and Brier scores declined relative to the baseline, while ECE values rose above 0.29 in every Bayesian run.

These results suggest that the VBLL-enhanced models are capable of improving predictive performance, particularly in high-signal datasets like Cleveland, but they may compromise calibration if not carefully tuned. Configurations C4 and C5, which used a higher $\mu$ coefficient and alternative KL annealing strategies (Table~\ref{tab:vbll-configs}), consistently performed better in terms of accuracy without drastic losses in AUC. However, no configuration managed to outperform the baseline in calibration metrics, indicating room for future improvement in uncertainty modeling, perhaps through temperature scaling or more adaptive priors. Under-performance of VBLL integrated TabPFN models compared to baseline was visually observerd from the reliability diagrams of each configuration against baseline see Appendix A, Fig. \ref{fig:reliability_grid}. 

In summary, although Variational Bayesian last layers present a hopeful path for uncertainty-aware learning, particularly within medical datasets, striking a balance between predictive accuracy and calibration continues to be a significant challenge.

\section{Conclusion}
This study examined the effects of integrating Variational Bayesian Last Layers (VBLL) into the TabPFN foundation model for uncertainty-aware learning in medical datasets. While most VBLL configurations increased accuracy and AUC, particularly on the Cleveland dataset, they generally led to inferior calibration metrics compared to the baseline.  These findings imply that, even though  Bayesian modeling improving uncertainty representation, it may also introduce overconfidence if not properly tuned.

It's interesting to note that, particularly on the PIMA dataset, the TabPFN baseline, without any explicit Bayesian extensions, performed well in terms of calibration and predictive accuracy. This indicates that, its thorough pretraining lead the underlying foundation model to already capture valuable uncertainty representations. As a result, it is evident that VBLL layers may not always give additional advantage and may even be harmful when improperly configured.

Future work can explore more adaptive Bayesian integration methods or can leverage post-hoc calibration techniques such as temperature scaling. Furthermore, understanding the circumstances under which foundation models suitable for uncertainty estimation remains an important path for further research.

\bibliographystyle{IEEEtran}
\bibliography{research}

\appendices
\section{Reliability Diagrams Across Configurations}
\begin{figure*}[t]
    \centering
    \includegraphics[width=\textwidth]{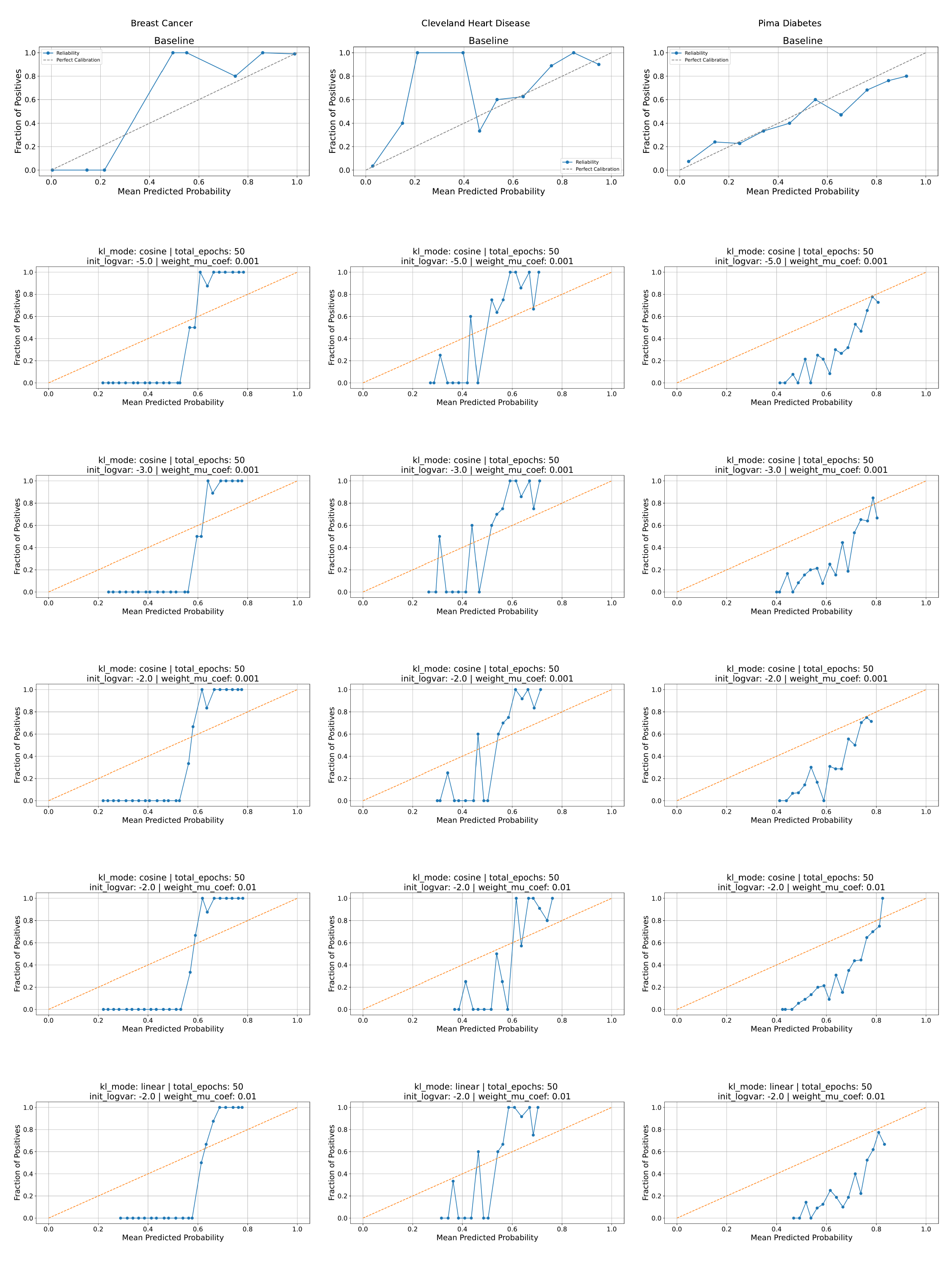}
    \caption{Reliability diagrams for all datasets across VBLL configurations and the baseline. Each row corresponds to a different VBLL hyperparameter configuration, and each column corresponds to a different dataset.}
    \label{fig:reliability_grid}
\end{figure*}

\end{document}